\documentclass[10pt,twocolumn,letterpaper]{article}

\usepackage{iccv}
\usepackage{times}
\usepackage{epsfig}
\usepackage{graphicx}
\usepackage{amsmath}
\usepackage{amssymb}

\usepackage{multirow}
\usepackage{caption}
\usepackage{subcaption}
\usepackage{algorithm}
\usepackage{algpseudocode}
\usepackage{float}

\usepackage{color}
\usepackage{framed}
\definecolor{shadecolor}{rgb}{0.9,0.9,0.9}

\usepackage[pagebackref=true,breaklinks=true,letterpaper=true,colorlinks,bookmarks=false]{hyperref}

\newcommand{\by} {{\bf y }}

\newcommand{\bk} {{\bf k }}

\newcommand{\bo} {{\bf o }}

\newcommand{\bI} {{\bf I }}

\newcommand{\vgg} {VGG$_{16}$}

\newcommand{\tabincell}[2]{\begin{tabular}{@{}#1@{}}#2\end{tabular}}

\iccvfinalcopy 

\def\iccvPaperID{901} 

\ificcvfinal\fi
\begin{document}

\title{Semantic Image Segmentation via Deep Parsing Network\thanks{This work has been accepted to appear in ICCV 2015. This is the pre-printed version. Content may slightly change prior to the final publication.}}

\author{Ziwei Liu$^{\dag}$ \hspace{8pt} Xiaoxiao Li\thanks{indicates shared first authorship.} \hspace{8pt} Ping Luo \hspace{8pt} Chen Change Loy \hspace{8pt} Xiaoou Tang\\
Department of Information Engineering, The Chinese University of Hong Kong\\
{\tt\small \{lz013,lx015,pluo,ccloy,xtang\}@ie.cuhk.edu.hk}
}

\maketitle
\thispagestyle{empty}

\begin{abstract}

    This paper addresses
    semantic image segmentation
    by incorporating rich information into Markov Random Field (MRF), including high-order relations and mixture of label contexts.
    Unlike previous works that optimized MRFs using iterative algorithm, we solve MRF by proposing a Convolutional Neural Network (CNN), namely Deep Parsing Network (DPN), which enables deterministic end-to-end computation in a single forward pass.
    Specifically, DPN extends a contemporary CNN architecture to model unary terms and additional layers are carefully devised to approximate the mean field algorithm (MF) for pairwise terms.
    It has several appealing properties.
    First, different from the recent works that combined CNN and MRF, where many iterations of MF were required for each training image during back-propagation, DPN is able to achieve high performance by approximating one iteration of MF.
    %
    Second, DPN represents various types of pairwise terms, making many existing works as its special cases.
    Third, DPN makes MF easier to be parallelized and speeded up in Graphical Processing Unit (GPU).
    %
    %
    DPN is thoroughly evaluated on the PASCAL VOC 2012 dataset, where a single DPN model yields a new state-of-the-art segmentation accuracy of 77.5\%.

\end{abstract}



\section{Introduction}


Markov Random Field (MRF) or Conditional Random Field (CRF) has achieved great successes in semantic image segmentation, which is one of the most challenging problems in computer vision.
Existing works such as \cite{shi2000normalized, ren2003learning, felzenszwalb2006efficient, szummer2008learning, fulkerson2009class, arbelaez2011contour, farabet2013learning, mostajabi2014feedforward, long2014fully} can be generally categorized into two groups based on their definitions of the unary and pairwise terms of MRF.

In the first group, researchers improved labeling accuracy by exploring rich information to define the pairwise functions, including long-range dependencies \cite{koltun2011efficient, krahenbuhl2013parameter}, high-order potentials \cite{vineet2012filter, vineet2013posefield}, and semantic label contexts \cite{liu2011nonparametric, mottaghi2014role, yang2014context}.
For example, Kr{\"a}henb{\"u}hl \etal \cite{koltun2011efficient} attained accurate segmentation boundary by inferring on a fully-connected graph.
Vineet \etal \cite{vineet2012filter} extended \cite{koltun2011efficient} by defining both high-order and long-range terms between pixels.
Global or local semantic contexts between labels were also investigated by \cite{yang2014context}.
Although they accomplished promising results,
they modeled the unary terms as SVM or Adaboost, whose learning capacity becomes a bottleneck.
The learning and inference of complex pairwise terms are often expensive.

In the second group, people learned a strong unary classifier by leveraging the recent advances of deep learning, such as the Convolutional Neural Network (CNN).
With deep models, these works \cite{luo2012hierarchical, luo2013pedestrian, mostajabi2014feedforward, long2014fully, chen2014semantic, papandreou2015weakly, zheng2015conditional, schwing2015fully, lin2015efficient} demonstrated encouraging results using simple definition of the pairwise function or even ignore it.
%
For instance, Long \etal \cite{long2014fully} transformed fully-connected layers of CNN into convolutional layers, making accurate per-pixel classification possible using the contemporary CNN architectures that were pre-trained on ImageNet \cite{deng2009imagenet}.
Chen \etal \cite{chen2014semantic} improved \cite{long2014fully} by feeding the outputs of CNN into a MRF with simple pairwise potentials,
but it treated CNN and MRF as separated components.
A recent advance was obtained by \cite{schwing2015fully}, which jointly trained CNN and MRF by passing the error of MRF inference backward into CNN,
but iterative inference of MRF such as the mean field algorithm (MF) \cite{opper2001naive} is required for each training image during back-propagation (BP).
Zheng \etal \cite{zheng2015conditional} further showed that the procedure of MF inference can be represented as a Recurrent Neural Network (RNN), but their computational costs are similar.
We found that directly combing CNN and MRF as above is inefficient, because CNN typically has millions of parameters while MRF infers thousands of latent variables;
and even worse, incorporating complex pairwise terms into MRF becomes impractical,
limiting the performance of the entire system.

This work proposes a novel Deep Parsing Network (DPN),
which is able to jointly train CNN and complex pairwise terms.
%
%
DPN has several appealing \textbf{properties}.
(1) DPN solves MRF with a single feed-forward pass, reducing computational cost and meanwhile maintaining high performance.
Specifically, DPN models unary terms by extending the VGG-16 network (\vgg) \cite{simonyan2014very} pre-trained on ImageNet, while additional layers are carefully designed to model complex pairwise terms.
%
Learning of these terms is transformed into deterministic end-to-end computation by BP,
%
instead of embedding MF into BP as \cite{schwing2015fully, lin2015efficient} did.
Although MF can be represented by RNN \cite{zheng2015conditional}, it needs to recurrently compute the forward pass so as to achieve good performance and thus is time-consuming,
\eg each forward pass contains hundred thousands of weights.
%
DPN approximates MF by using only one iteration.
This is made possible by joint learning strong unary terms and rich pairwise information.
%
%
%
(2) Pairwise terms determine the graphical structure.
In previous works, if the former is changed, so is the latter as well as its inference procedure.
But with DPN, modifying the complexity of pairwise terms, \eg range of pixels and contexts, is as simple as modifying the receptive fields of convolutions, without varying BP.
DPN is able to represent multiple types of pairwise terms, making many previous works \cite{chen2014semantic, zheng2015conditional, schwing2015fully} as its special cases.
(3) 
DPN approximates MF with convolutional and pooling operations, which
can be speeded up by low-rank approximation \cite{jaderberg2014speeding} and easily parallelized \cite{chetlur2014cudnn} in a Graphical Processing Unit (GPU).

%

Our \textbf{contributions} are summarized as below. (1) A novel DPN is proposed to jointly train \vgg~and rich pairwise information, \ie \emph{mixture of label contexts} and \emph{high-order relations}.
Compared to existing deep models, DPN can approximate MF with only \emph{one iteration}, reducing computational cost but still maintaining high performance.
(2) We disclose that DPN represents multiple types of MRFs, making many previous works such as RNN \cite{zheng2015conditional} and DeepLab \cite{chen2014semantic} as its special cases.
(3) Extensive experiments investigate which component of DPN is crucial to achieve high performance.
%
%
%
A single DPN model achieves a new state-of-the-art accuracy of 77.5\% on the PASCAL VOC 2012 \cite{everingham2010pascal} test set.
(4) We analyze the time complexity of DPN on GPU.



\section{Our Approach}

DPN learns MRF by extending \vgg~to model unary terms and additional layers are carefully designed for pairwise terms.

\textbf{Overview~} 
MRF \cite{freeman2000learning} is an undirected graph where each node represents a pixel in an image $\bI$, and each edge represents relation between pixels.
Each node is associated with a binary latent variable, $y^i_u\in\{0,1\}$, indicating whether a pixel $i$ has label $u$.
We have $\forall u\in L=\{1,2,...,l\}$, representing a set of $l$ labels.
The energy function of MRF is written as
\begin{equation}\label{eq:E}
E(\by)=\sum_{\forall i\in\mathcal{V}}\Phi(y_i^u)+\sum_{\forall i,j\in\mathcal{E}}\Psi(y^u_i,y^v_j),
\end{equation}
where $\by$, $\mathcal{V}$, and $\mathcal{E}$ denote a set of latent variables, nodes, and edges, respectively.
$\Phi(y^u_i)$ is the unary term, measuring the cost of assigning label $u$ to the $i$-th pixel.
For instance, if pixel $i$ belongs to the first category other than the second one, we should have $\Phi(y^1_i)<\Phi(y_i^2)$.
Moreover, $\Psi(y_i^u,y_j^v)$ is the pairwise term that measures the penalty of assigning labels $u,v$ to pixels $i,j$ respectively.
%
%

Intuitively, the unary terms represent per-pixel classifications, while the pairwise terms represent a set of smoothness constraints.
The unary term in Eqn.(\ref{eq:E}) is typically defined as
\begin{equation}\label{eq:unary}
\Phi(y_i^u)=-\ln p(y_i^u=1|\bI)
\end{equation}
where $p(y_i^u=1|\bI)$ indicates the probability of the presence of label $u$ at pixel $i$, modeling by \vgg.
To simplify discussions, we abbreviate it as $p^u_i$.
%
%
The smoothness term can be formulated as
\begin{equation}\label{eq:old_p}
\Psi(y_i^u,y_j^v)=\mu(u,v)\mathrm{d}(i,j),
\end{equation}
where
the first term learns the penalty of global co-occurrence between any pair of labels, \eg the output value of $\mu(u,v)$ is large if $u$ and $v$ should not coexist,
while the second term calculates the distances between pixels, \eg
$\mathrm{d}(i,j)=\omega_1\|\bI_i-\bI_j\|^2+\omega_2\|[x_i~y_i]-[x_j~y_j]\|^2$.
Here, $\bI_i$ indicates a feature vector such as RGB values extracted from the $i$-th pixel, $x,y$ denote coordinates of pixels' positions, and $\omega_1,\omega_2$ are the constant weights.
Eqn.(\ref{eq:old_p}) implies that if two pixels are close and look similar, they are encouraged to have labels that are compatible.
It has been adopted by most of the recent deep models \cite{chen2014semantic, zheng2015conditional, schwing2015fully} for semantic image segmentation.

However, Eqn.(\ref{eq:old_p}) has two main drawbacks.
First, its first term captures the co-occurrence frequency of two labels in the training data, but neglects the spatial context between objects.
For example, `person' may appear beside `table', but not at its bottom.
This spatial context is a mixture of patterns, as different object configurations may appear in different images.
Second, it defines only the pairwise relations between pixels, missing their high-order interactions.

\begin{figure}[t]
  \centering
  \includegraphics[width=0.37\textwidth]{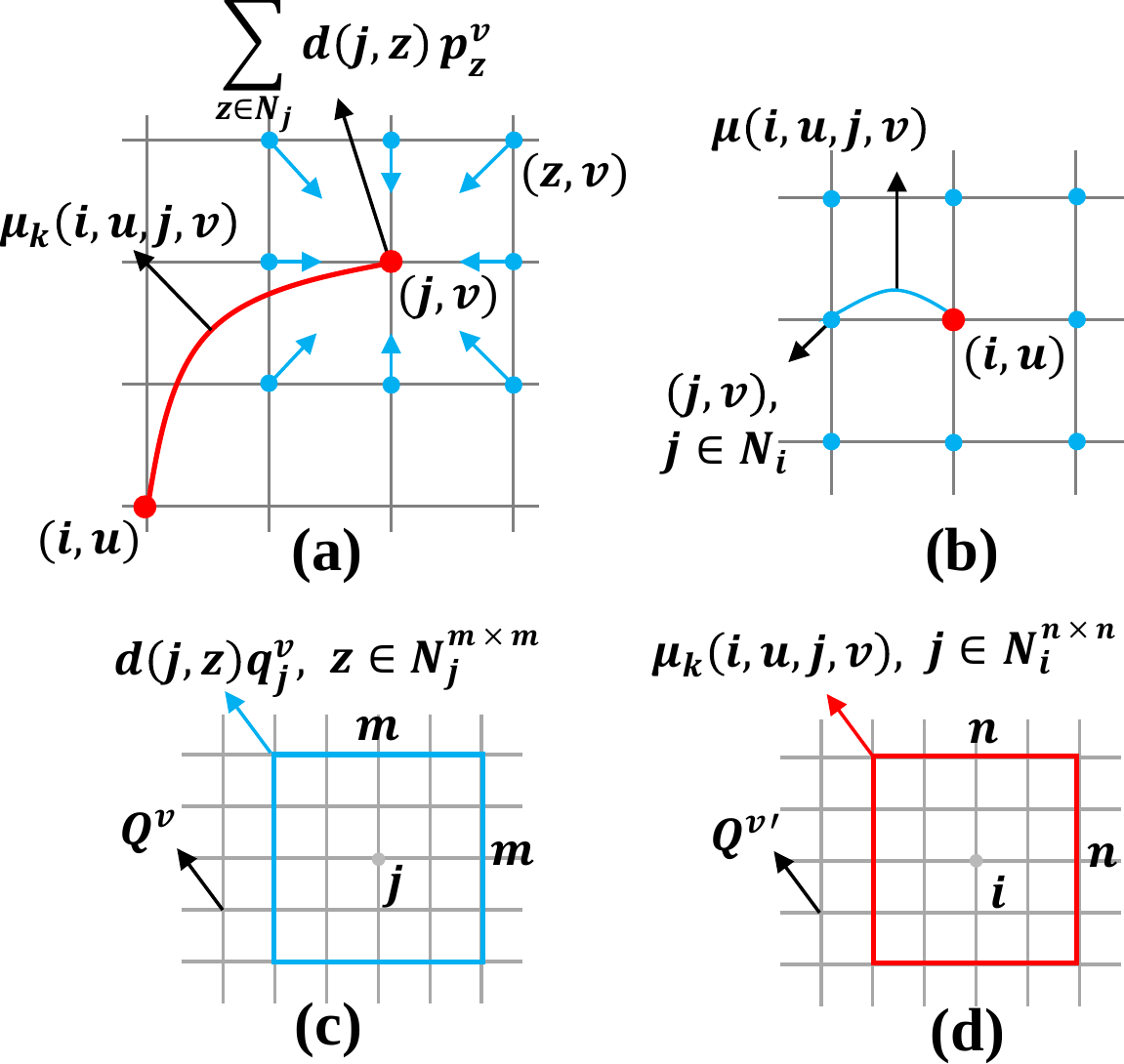}
  \caption{\footnotesize (a) Illustration of the pairwise terms in DPN. (b) explains the label contexts. (c) and (d) show that mean field update of DPN corresponds to convolutions.}
  \label{fig:mrf}
  \vspace{-8pt}
\end{figure}

To resolve these issues, we define the smoothness term by leveraging rich information between pixels, which is one of the \textbf{advantages} of DPN over existing deep models.
%
We have
\begin{equation}\label{eq:pairwise}
\Psi(y_i^u,y_j^v)=\sum_{k=1}^K\lambda_k\mu_k(i,u,j,v)
\sum_{\forall z\in \mathcal{N}_j}\mathrm{d}(j,z)p^v_z.
\end{equation}
The first term in Eqn.(\ref{eq:pairwise})
%
learns a \textbf{mixture of local label contexts}, penalizing label assignment in a local region,
where $K$ is the number of components in mixture and $\lambda_k$ is an indicator, determining which component is activated. We define $\lambda_k\in\{0,1\}$ and $\sum_{k=1}^K\lambda_k=1$.
%
An intuitive illustration is given in Fig.\ref{fig:mrf} (b),
where the dots in red and blue represent a center pixel $i$ and its neighboring pixels $j$, \ie $j\in\mathcal{N}_i$, and $(i,u)$ indicates assigning label $u$ to pixel $i$.
Here, $\mu(i,u,j,v)$ outputs labeling cost between $(i,u)$ and $(j,v)$ with respect to their relative positions.
For instance, if $u,v$ represent `person' and `table', the learned penalties of positions $j$ that are at the bottom of center $i$ should be large.
%
%
The second term basically models a \textbf{triple penalty}, which involves pixels $i$, $j$, and $j$'s neighbors,
implying that if $(i,u)$ and $(j,v)$ are compatible, then $(i,u)$ should be also compatible with $j$'s nearby pixels $(z,v)$, $\forall z\in\mathcal{N}_j$, as shown in Fig.\ref{fig:mrf} (a).

Learning parameters (\ie weights of \vgg~and costs of label contexts) in Eqn.(\ref{eq:E}) is to
%
minimize the distances between ground-truth label map and $\by$, which needs to be inferred subject to the smoothness constraints.

\textbf{Inference Overview}
Inference of Eqn.(\ref{eq:E}) can be obtained by the mean field (MF) algorithm \cite{opper2001naive}, which
estimates the joint distribution of MRF, $P(\by)$$=$$\frac{1}{Z}\exp\{-E(\by)\}$, by using a fully-factorized proposal distribution, $Q(\by)$ $=$ $\prod_{\forall i\in\mathcal{V}}$$\prod_{\forall u\in L}$$q^u_i$,
where each $q_i^u$ is a variable we need to estimate, indicating the predicted probability of assigning label $u$ to pixel $i$.
To simplify the discussion, we denote $\Phi(y_i^u)$ and $\Psi(y_i^u,y_j^v)$ as $\Phi_i^u$ and $\Psi_{ij}^{uv}$, respectively.
$Q(\by)$ is typically optimized by minimizing a free energy function \cite{jordan1999introduction} of MRF,
\begin{eqnarray}\label{eq:FQ}
F(Q)&=&\sum_{\forall i\in\mathcal{V}}\sum_{\forall u\in L}q_i^u\Phi_i^u+\sum_{\forall i,j\in\mathcal{E}}\sum_{\forall u\in L}\sum_{\forall v\in L}q_i^u q_j^v \Psi_{ij}^{uv}\nonumber\\
&&+\sum_{\forall i\in\mathcal{V}}\sum_{\forall u\in L} q_i^u\ln q_i^u.
\end{eqnarray}
Specifically, the first term in Eqn.(\ref{eq:FQ}) characterizes the cost of each pixel's predictions, while the second term characterizes the consistencies of predictions between pixels. The last term is the entropy, measuring the confidences of predictions.
To estimate $q_i^u$, we differentiate Eqn.(\ref{eq:FQ}) with respect to it and equate the resulting expression to zero. We then have a closed-form expression,
%
\begin{equation}\label{eq:q}
q_i^u\propto\exp\big\{-(\Phi_i^u+\sum_{\forall j\in\mathcal{N}_i}\sum_{\forall v\in L}q_j^v\Psi_{ij}^{uv})\big\},
\end{equation}
%
such that the predictions for each pixel is independently attained by repeating Eqn.(\ref{eq:q}),
which implies whether pixel $i$ have label $u$ is proportional to the estimated probabilities of all its neighboring pixels, weighted by their corresponding smoothness penalties.
Substituting Eqn.(\ref{eq:pairwise}) into (\ref{eq:q}), we have
\begin{eqnarray}\label{eq:newq}
q_i^u&\propto& \exp\Big\{
-\Phi_i^u-\sum_{k=1}^K\lambda_k\sum_{\forall v\in L}\sum_{\forall j\in\mathcal{N}_i}\\\nonumber
&&\mu_k(i,u,j,v)
\sum_{\forall z\in\mathcal{N}_j}\mathrm{d}(j,z)q_j^v q^v_z
\Big\},
\end{eqnarray}
where each $q^u_i$ is initialized by the corresponding $p^u_i$ in Eqn.(\ref{eq:unary}), which is the unary prediction of \vgg.
Eqn.(\ref{eq:newq}) satisfies the smoothness constraints.

In the following, DPN approximates one iteration of Eqn.(\ref{eq:newq}) by decomposing it into two steps.
%
Let $Q^v$ be a predicted label map of the $v$-th category.
In the first step as shown in Fig.\ref{fig:mrf} (c),
%
we calculate the triple penalty term in (\ref{eq:newq}) by applying a $m\times m$ filter on each position $j$, where
each element of this filter equals $\mathrm{d}(j,z)q^v_j$, resulting in ${Q^v}'$.
%
%
Apparently, this step smoothes the prediction of pixel $j$ with respect to the distances between it and its neighborhood.
In the second step as illustrated in (d),
the labeling contexts can be obtained by convolving ${Q^v}'$ with a $n\times n$ filter, each element of which equals $\mu_k(i,u,j,v)$, penalizing the triple relations as shown in (a).

\section{Deep Parsing Network}

\begin{table*}[t]
\scriptsize
\begin{center}
\begin{tabular}{c|c|c|c|c|c|c|c|c|c|c|c|c|c|c|c}
\multicolumn{16}{c}{(a)~\textbf{\vgg:}~~224$\times$224$\times$3 \emph{input image};~~1$\times$1000 \emph{output labels}} \\
\hline
\multicolumn{1}{c}{} & \multicolumn{1}{c}{1} & \multicolumn{1}{c}{2} & \multicolumn{1}{c}{3} & \multicolumn{1}{c}{4} & \multicolumn{1}{c}{5} & \multicolumn{1}{c}{6} & \multicolumn{1}{c}{7} & \multicolumn{1}{c}{8} & \multicolumn{1}{c}{9} & \multicolumn{1}{c}{10} & \multicolumn{1}{c}{11} &
\multicolumn{1}{c}{12} &\multicolumn{1}{c}{} & \multicolumn{1}{c}{} & \multicolumn{1}{c}{}\\
\hline
\tabincell{r}{\emph{\textbf{layer}}\\\emph{\textbf{filter--stride}}\\\emph{\textbf{\#channel}}\\\emph{\textbf{activation}}\\\emph{\textbf{size}}} &
\tabincell{c}{2$\times$\textbf{conv}\\3-1\\64\\ $\mathrm{relu}$\\ 224} & 
\tabincell{c}{\textbf{max}\\2-2\\64\\ $\mathrm{idn}$\\112} &
\tabincell{c}{2$\times$\textbf{conv}\\3-1\\128\\$\mathrm{relu}$\\ 112} &
\tabincell{c}{\textbf{max}\\2-2\\128\\ $\mathrm{idn}$\\56} &
\tabincell{c}{3$\times$\textbf{conv}\\3-1\\256 \\$\mathrm{relu}$\\ 56} &
\tabincell{c}{\textbf{max}\\2-2\\256\\ $\mathrm{idn}$\\28} &
\tabincell{c}{3$\times$\textbf{conv}\\3-1\\ 512 \\$\mathrm{relu}$\\ 28} &
\tabincell{c}{\textbf{max}\\2-2\\512\\ $\mathrm{idn}$\\14} &
\tabincell{c}{3$\times$\textbf{conv}\\3-1\\ 512 \\$\mathrm{relu}$\\ 14} &
\tabincell{c}{\textbf{max}\\2-2\\512\\ $\mathrm{idn}$\\7} &
\tabincell{c}{2$\times$\textbf{fc}\\-\\1\\ $\mathrm{relu}$\\4096} &
\tabincell{c}{\textbf{fc}\\-\\1\\ $\mathrm{soft}$\\1000} &
\multicolumn{1}{c}{} &
\multicolumn{1}{c}{} & \multicolumn{1}{c}{}
\\
\hline
\multicolumn{16}{c}{(b)~\textbf{DPN:}~~512$\times$512$\times$3 \emph{input image};~~512$\times$512$\times$21 \emph{output label maps}} \\
\hline
\multicolumn{1}{c}{} & \multicolumn{1}{c}{1} & \multicolumn{1}{c}{2} & \multicolumn{1}{c}{3} & \multicolumn{1}{c}{4} & \multicolumn{1}{c}{5} & \multicolumn{1}{c}{6} & \multicolumn{1}{c}{7} & \multicolumn{1}{c}{8} & \multicolumn{1}{c}{9} & \multicolumn{1}{c}{10} & \multicolumn{1}{c}{11} &
\multicolumn{1}{c}{12} & \multicolumn{1}{c}{13} & \multicolumn{1}{c}{14} &
\multicolumn{1}{c}{15} \\
\hline
\tabincell{r}{\emph{\textbf{layer}}\\\emph{\textbf{filter--stride}}\\\emph{\textbf{\#channel}}\\\emph{\textbf{activation}}\\\emph{\textbf{size}}} &
\tabincell{c}{2$\times$\textbf{conv}\\3-1\\64\\$\mathrm{relu}$\\ 512} &
\tabincell{c}{\textbf{max}\\2-2\\64\\$\mathrm{idn}$\\ 256} &
\tabincell{c}{2$\times$\textbf{conv}\\3-1\\128\\ $\mathrm{relu}$\\256} &
\tabincell{c}{\textbf{max}\\2-2\\128\\ $\mathrm{idn}$\\128} &
\tabincell{c}{3$\times$\textbf{conv}\\3-1\\256 \\$\mathrm{relu}$\\ 128} &
\tabincell{c}{\textbf{max}\\2-2\\ 256\\ $\mathrm{idn}$\\64} &
\tabincell{c}{3$\times$\textbf{conv}\\3-1\\ 512 \\ $\mathrm{relu}$\\64} &
\tabincell{c}{3$\times$\textbf{conv}\\5-1\\ 512 \\$\mathrm{relu}$\\ 64} &
\tabincell{c}{\textbf{conv}\\25-1\\ 4096 \\$\mathrm{relu}$\\ 64} &
\tabincell{c}{\textbf{conv}\\1-1\\ 4096 \\$\mathrm{relu}$\\ 64} &
\tabincell{c}{\textbf{conv}\\1-1\\ 21 \\$\mathrm{sigm}$\\ 512} &
\tabincell{c}{\textbf{lconv}\\50-1\\21\\ $\mathrm{lin}$\\512} &
\tabincell{c}{\textbf{conv}\\9-1\\105\\ $\mathrm{lin}$\\512} &
\tabincell{c}{\textbf{bmin} \\ 1-1 \\ 21 \\ $\mathrm{idn}$\\512} &
\tabincell{c}{\textbf{sum} \\ 1-1 \\ 21 \\ $\mathrm{soft}$\\512}\\
\hline
\end{tabular}
\end{center}
\vspace{-10pt}
\caption{\footnotesize The comparisons between the network architectures of \vgg~and DPN, as shown in (a) and (b) respectively. Each table contains five rows, representing the `\textbf{name of layer}', `\textbf{receptive field of filter}'$-$`\textbf{stride}', `\textbf{number of output feature maps}', `\textbf{activation function}' and `\textbf{size of output feature maps}', respectively. Furthermore, `\textbf{conv}', `\textbf{lconv}',`\textbf{max}', `\textbf{bmin}', `\textbf{fc}', and `\textbf{sum}' represent the convolution, local convolution, max pooling, block min pooling, fully connection, and summation, respectively. Moreover, `relu', `idn', `soft', `sigm', and `lin' represent the activation functions, including rectified linear unit \cite{krizhevsky2012imagenet}, identity, softmax, sigmoid, and linear, respectively.}
\label{tab:net}
\end{table*}

%
This section describes the implementation of Eq.(\ref{eq:newq}) in a Deep Parsing Network (DPN).
DPN extends \vgg~as unary term and additional layers are designed to approximate one iteration of MF inference as the pairwise term.
%
%
The hyper-parameters of \vgg~and DPN are compared in Table \ref{tab:net}.

\textbf{\vgg~}
As listed in Table \ref{tab:net} (a), the first row represents the \emph{name} of layer
and `$x$-$y$' in the second row represents the \emph{size} of the receptive field and the \emph{stride} of convolution, respectively.
For instance, `3-1' in the convolutional layer implies that the receptive field of each filter is 3$\times$3 and it is applied on every single pixel of an input feature map, while `2-2' in the max-pooling layer indicates each feature map is pooled over every other pixel within a 2$\times$2 local region.
The last three rows show the number of the output feature maps, activation functions, and the size of output feature maps, respectively.
As summarized in Table \ref{tab:net} (a), \vgg~contains thirteen convolutional layers, five max-pooling layers, and three fully-connected layers.
These layers can be partitioned into twelve groups, each of which covers one or more homogenous layers.
For example, the first group comprises two convolutional layers with 3$\times$3 receptive field and 64 output feature maps, each of which is 224$\times$224.

\subsection{Modeling Unary Terms}

To make full use of \vgg, which is pre-trained by ImageNet,
we adopt all its parameters to initialize the filters of the first ten groups of DPN. To simplify the discussions, we take PASCAL VOC 2012 (VOC12) \cite{everingham2010pascal} as an example. Note that DPN can be easily adapted to any other semantic image segmentation dataset by modifying its hyper-parameters.
VOC12 contains 21 categories and each image is rescaled to 512$\times$512 in training. Therefore, DPN needs to predict totally 512$\times$512$\times$21 labels, \ie one label for each pixel.
To this end, we
extends \vgg~in two aspects.

\begin{figure}[t]
  \centering
  \includegraphics[width=0.4\textwidth]{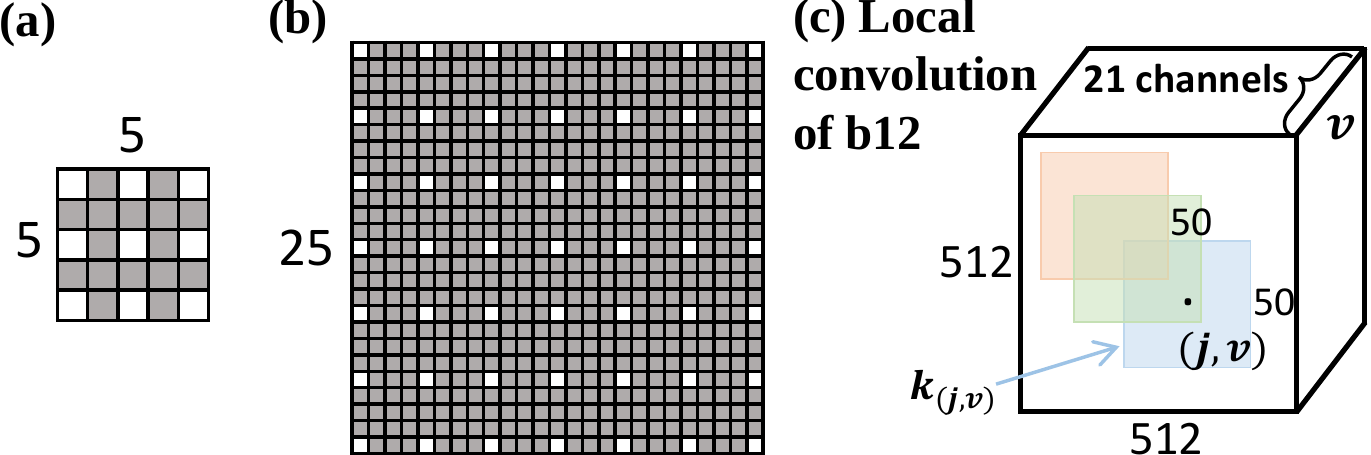}
  \caption{\footnotesize (a) and (b) show the padding of the filters. (c) illustrates local convolution of b12.}
  \label{fig:filter}
  \vspace{-8pt}
\end{figure}

In particular, let a$i$ and b$i$ denote the $i$-th group in Table \ref{tab:net} (a) and (b), respectively.
First, we \textbf{increase resolution} of \vgg~by removing its max pooling layers at a8 and a10, because most of the information is lost after pooling, \eg a10 reduces the input size by 32 times, \ie from 224$\times$224 to 7$\times$7.
As a result, the smallest size of feature map in DPN is 64$\times$64, keeping much more information compared with \vgg.
Note that the filters of b8 are initialized as the filters of a9, but the 3$\times$3 receptive field is padded into 5$\times$5 as shown in Fig.\ref{fig:filter} (a), where the cells in white are the original values of the a9's filter and the cells in gray are zeros.
This is done because a8 is not presented in DPN, such that each filter in a9 should be convolved on every other pixel of a7.
To maintain the convolution with one stride, we pad the filters with zeros.
Furthermore, the feature maps in b11 are up-sampled to 512$\times$512 by bilinear interpolation.
Since DPN is trained with label maps of the entire images, the missing information in the preceding layers of b11 can be recovered by BP.

Second, two fully-connected layers at a11 are transformed to two convolutional layers at b9 and b10, respectively.
As shown in Table \ref{tab:net} (a), the first `fc' layer learns 7$\times$7$\times$512$\times$4096 parameters, which can be altered to 4096 filters in b9, each of which is 25$\times$25$\times$512.
%
Since a8 and a10 have been removed, the 7$\times$7 receptive field is padded into 25$\times$25 similar as above and shown in Fig.\ref{fig:filter} (b).
The second `fc' layer learns a 4096$\times$4096 weight matrix, corresponding to 4096 filters in b10. Each filter is 1$\times$1$\times$4096.

Overall, b11 generates the unary labeling results, producing twenty-one 512$\times$512 feature maps, each of which represents the probabilistic label map of each category.

\subsection{Modeling Smoothness Terms}
The last four layers of DPN, \ie from b12 to b15, are carefully designed to smooth the unary labeling results.

$\bullet$ \textbf{b12} As listed in Table \ref{tab:net} (b), `lconv' in b12 indicates a \textbf{locally convolutional layer}, which is widely used in face recognition \cite{sun2014deep, taigman2014deepface} to capture different information from different facial positions.
Similarly, distinct spatial positions of b12 have different filters, and each filter is shared across 21 input channels, as shown in Fig.\ref{fig:filter} (c).
It can be formulated as
\begin{equation}\label{eq:lconv}
\bo^{12}_{(j,v)}=\mathrm{lin}(\bk_{(j,v)}\ast\bo^{11}_{(j,v)}),
\end{equation}
where $\mathrm{lin}(x)=ax+b$ representing the linear activation function, `$\ast$' is the convolutional operator, and $\bk_{(j,v)}$ is a 50$\times$50$\times$1 filter at position $j$ of channel $v$.
We have $\bk_{(j,1)}=\bk_{(j,2)}=...=\bk_{(j,21)}$ shared across 21 channels.
$\bo^{11}_{(j,v)}$ indicates a local patch in b11, while $\bo^{12}_{(j,v)}$ is the corresponding output of b12.
%
%
Since b12 has stride one, the result of $\bk_{j}\ast\bo^{11}_{(j,v)}$ is a scalar.
In summary, b12 has 512$\times$512 different filters and produces 21 output feature maps.

Eqn.(\ref{eq:lconv}) implements the \textbf{triple penalty} of Eqn.(\ref{eq:newq}).
Recall that each output feature map of b11 indicates a probabilistic label map of a specific object appearing in the image.
As a result, Eqn.(\ref{eq:lconv}) suggests that the probability of object $v$ presented at position $j$ is updated by weighted averaging over the probabilities at its nearby positions.
Thus, as shown in Fig.\ref{fig:mrf} (c), $\bo^{11}_{(j,v)}$ corresponds to a patch of $Q^v$ centered at $j$, which has values $p^v_z$, $\forall z\in\mathcal{N}_j^{50\times50}$.
Similarly, $\bk_{(j,v)}$ is initialized by $\mathrm{d}(j,z)p^v_j$, implying each filter captures dissimilarities between positions.
These filters remain fixed during BP,
other than learned as in conventional CNN\footnote{Each filter in b12 actually represents a distance metric between pixels in a specific region.
In VOC12, the patterns of all the training images in a specific region are heterogenous, because of various object shapes.
Therefore, we initialize each filter with Euclidean distance.
Nevertheless, Eqn.(\ref{eq:lconv}) is a more general form than the triple penalty in Eqn.(\ref{eq:newq}), \ie filters in (\ref{eq:lconv}) can be automatically learned from data,
if the patterns in a specific region are homogenous, such as face or human images, which have more regular shapes than images in VOC12.}.

\begin{figure}[t]
  \centering
  \includegraphics[width=0.4\textwidth]{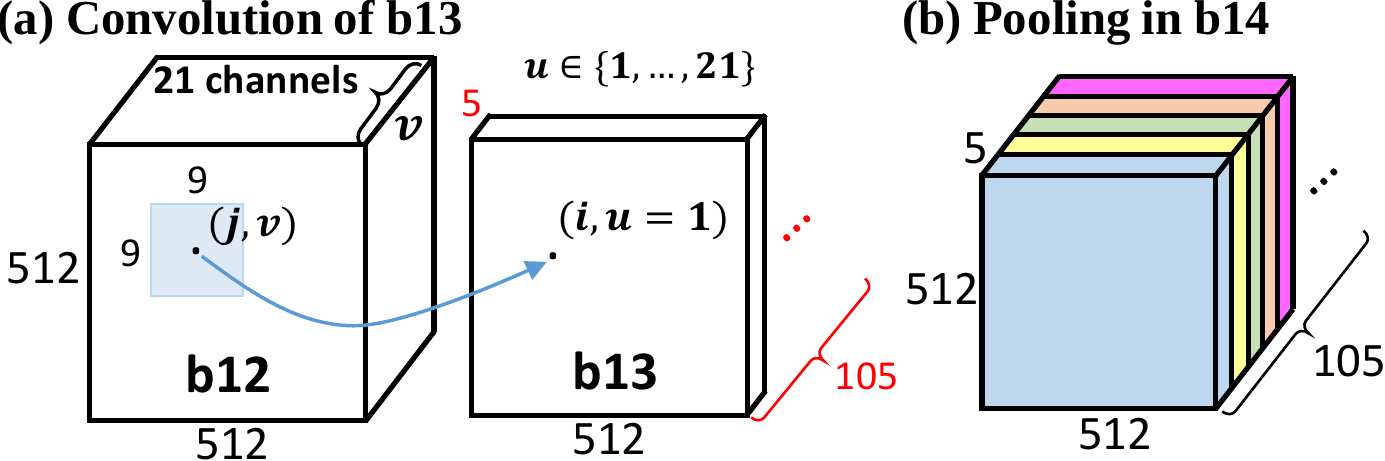}
  \caption{\footnotesize (a) and (b) illustrates the convolutions of b13 and the poolings in b14.}
  \label{fig:b1314}
  \vspace{-8pt}
\end{figure}

$\bullet$ \textbf{b13}
As shown in Table \ref{tab:net} (b) and Fig.\ref{fig:b1314} (a), b13 is a convolutional layer that generates 105 feature maps by using 105 filters of size 9$\times$9$\times$21.
For example, the value of $(i,u=1)$ is attained by applying a 9$\times$9$\times$21 filter at positions $\{(j,v=1,...,21)\}$.
In other words, b13 learns a filter for each category to penalize the probabilistic label maps of b12, corresponding to the \textbf{local label contexts} in Eqn.(\ref{eq:newq}) by assuming $K=5$ and $n=9$, as shown in Fig.\ref{fig:mrf} (d).
%

%
%

$\bullet$ \textbf{b14} As illustrated in Table \ref{tab:net} and Fig.\ref{fig:b1314} (b), b14 is a block min pooling layer that pools over every 1$\times$1 region with one stride across every 5 input channels, leading to 21 output channels, \ie 105$\div$5$=$21.
b14 activates the contextual pattern with the smallest penalty.

$\bullet$ \textbf{b15} This layer combines both the unary and smoothness terms by summing the outputs of b11 and b14 in an element-wise manner similar to Eqn.(\ref{eq:newq}),
\begin{equation}
\bo^{15}_{(i,u)}=\frac{\exp\big\{\ln(\bo^{11}_{(i,u)})-\bo^{14}_{(i,u)}\big\}}
{\sum_{u=1}^{21}\exp\big\{
\ln(\bo^{11}_{(i,u)})-\bo^{14}_{(i,u)}\big\}},
\end{equation}
where probability of assigning label $u$ to pixel $i$ is normalized over all the labels.

\vspace{5pt}
\textbf{Relation to Previous Deep Models~}
Many existing deep models such as \cite{zheng2015conditional, chen2014semantic, schwing2015fully} employed Eqn.(\ref{eq:old_p}) as the pairwise terms,
which are the special cases of Eqn.(\ref{eq:newq}).
To see this, let $K$$=$1 and $j$$=$$i$, the right hand side of (\ref{eq:newq}) reduces to
%
%
%
\begin{eqnarray}\label{eq:K1s1}
&&\exp\{-\Phi_i^u-\sum_{v\in L}\lambda_1\mu_1(i,u,i,v)\sum_{z\in\mathcal{N}_i} \mathrm{d}(i,z)p_i^vp_z^v\}\nonumber\\
&=&\exp\{-\Phi_i^u-\sum_{v\in L}\mu(u,v)\sum_{z\in\mathcal{N}_i,z\neq i} \mathrm{d}(i,z)p_z^v\},
\end{eqnarray}
where $\mu(u,v)$ and $\mathrm{d}(i,z)$ represent the global label co-occurrence and pairwise pixel similarity of Eqn.(\ref{eq:old_p}), respectively.
This is because $\lambda_1$ is a constant, $\mathrm{d}(i,i)=0$, and $\mu(i,u,i,v)=\mu(u,v)$.
%
Eqn.(\ref{eq:K1s1}) is the corresponding MF update equation of (\ref{eq:old_p}).

\subsection{Learning Algorithms}\label{sec:learn}

\textbf{Learning} The first ten groups of DPN are initialized by \vgg\footnote{We use the released \vgg~model, which is public available at \url{http://www.robots.ox.ac.uk/~vgg/research/very_deep/}}, while the last four groups can be initialized randomly.
DPN is then fine-tuned in an incremental manner with four stages.
%
During fine-tuning, all these stages solve the pixelwise softmax loss \cite{long2014fully}, but updating different sets of parameters.

First, we add a loss function to b11 and fine-tune the weights from b1 to b11 without the last four groups, in order to learn the unary terms.
Second, to learn the triple relations, we stack b12 on top of b11 and update its parameters (\ie $\omega_1,\omega_2$ in the distance measure), but the weights of the preceding groups (\ie b1$\sim$b11) are fixed.
Third, b13 and b14 are stacked onto b12 and similarly, their weights are updated with all the preceding parameters fixed, so as to learn the local label contexts.
Finally, all the parameters are jointly fine-tuned.
%

\textbf{Implementation} DPN transforms Eqn.(\ref{eq:newq}) into convolutions and poolings in the groups from b12 to b15, such that filtering at each pixel can be performed in a parallel manner.
Assume we have $f$ input and $f'$ output feature maps, $N\times N$ pixels, filters with $s\times s$ receptive field, and a mini-batch with $M$ samples.
b12 takes a total $f\cdot N^2\cdot s^2\cdot M$ operations, b13 takes $f\cdot f'\cdot N^2\cdot s^2\cdot M$ operations, while both b14 and b15 require $f\cdot N^2\cdot M$ operations.
For example, when $M$$=$10 as in our experiment, we have 21$\times$512$^2$$\times$50$^2$$\times$10$=$1.3$\times$10$^{11}$ operations in b12, which has the highest complexity in DPN.
We parallelize these operations using matrix multiplication on GPU as \cite{chetlur2014cudnn} did, b12 can be computed within 30ms.
The total runtime of the last four layers of DPN is 75ms.
Note that convolutions in DPN can be further speeded up by low-rank decompositions \cite{jaderberg2014speeding} of the filters and model compressions \cite{hinton2014distilling}.

However, direct calculation of Eqn.(\ref{eq:newq}) is accelerated by fast Gaussian filtering \cite{adams2010fast}.
For a mini-batch of ten 512$\times$512 images, a recently optimized implementation \cite{koltun2011efficient} takes 12 seconds on CPU to compute one iteration of (\ref{eq:newq}).
%
Therefore, DPN makes (\ref{eq:newq}) easier to be parallelized and speeded up.
%


\section{Experiments}

\textbf{Dataset} We evaluate the proposed approach on the PASCAL VOC 2012 (VOC12) \cite{everingham2010pascal} dataset, which contains 20 object categories and one background category.
Following previous works such as \cite{hariharan2011semantic, long2014fully, chen2014semantic}, we employ $10,582$ images for training, $1,449$ images for validation, and $1,456$ images for testing.

\textbf{Evaluation Metrics}
All existing works employed mean pixelwise intersection-over-union (denoted as mIoU) \cite{long2014fully} to evaluate their performance.
To fully examine the effectiveness of DPN, we introduce another three metrics, including tagging accuracy (TA), localization accuracy (LA), and boundary accuracy (BA).
(1) TA compares the predicted image-level tags with the ground truth tags, calculating the accuracy of multi-class image classification.
%
%
(2) LA evaluates the IoU between the predicted object bounding boxes\footnote{They are the bounding boxes of the predicted segmentation regions.} and the ground truth bounding boxes (denoted as bIoU), measuring the precision of object localization.
(3) For those objects that have been correctly localized, we compare the predicted object boundary with the ground truth boundary,
measuring the precision of semantic boundary similar to \cite{hariharan2011semantic}.

\textbf{Comparisons}
%
%
DPN is compared with the best-performing methods
on VOC12, including FCN \cite{long2014fully}, Zoom-out \cite{mostajabi2014feedforward}, DeepLab \cite{chen2014semantic}, WSSL \cite{papandreou2015weakly}, BoxSup \cite{dai2015boxsup}, Piecewise \cite{lin2015efficient}, and RNN \cite{zheng2015conditional}.
All these methods are based on CNNs and MRFs, and trained on VOC12 data following \cite{long2014fully}.
They can be grouped according to different aspects:
(1) \textbf{joint-train}: Piecewise and RNN; (2) \textbf{w/o joint-train}: DeepLab, WSSL, FCN, and BoxSup; (3) \textbf{pre-train on COCO}: RNN, WSSL, and BoxSup.
The first and the second groups are the methods with and without joint training CNNs and MRFs, respectively.
%
%
Methods in the last group also employed MS-COCO \cite{lin2014microsoft} to pre-train deep models.
To conduct a comprehensive comparison, the performance of DPN are reported on both settings, \ie, with and without pre-training on COCO.

In the following, Sec.\ref{sec:ablation} investigates the effectiveness of different components of DPN on the VOC12 \emph{validation set}.
Sec.\ref{sec:overall} compares DPN with the state-of-the-art methods on the VOC12 \emph{test set}.

\begin{table}
\footnotesize

\begin{subtable}{\linewidth}
\vspace{4pt}
\centering
\begin{tabular}{c|c|c|c|c}
\hline
Receptive Field & baseline & 10$\times$10 & 50$\times$50 & 100$\times$100 \\
\hline\hline
mIoU (\%) & 63.4 & 63.8 & \textbf{64.7} & 64.3 \\
\hline
\end{tabular}\vspace{-5pt}
\subcaption{\footnotesize Comparisons between different receptive fields of b12.}
\end{subtable}

\begin{subtable}{\linewidth}
\vspace{4pt}
\centering
\begin{tabular}{c|c|c|c|c}
\hline
Receptive Field & 1$\times$1 & 5$\times$5 & 9$\times$9 & 9$\times$9 mixtures \\
\hline\hline
mIoU (\%) & 64.8 & 66.0 & 66.3 & \textbf{66.5} \\
\hline
\end{tabular}\vspace{-5pt}
\subcaption{\footnotesize Comparisons between different receptive fields of b13.}
\end{subtable}

\begin{subtable}{\linewidth}
\vspace{4pt}
\centering
\begin{tabular}{c|c|c|c}
\hline
Pairwise Terms & DSN~\cite{schwing2015fully} & DeepLab~\cite{chen2014semantic} & DPN \\
\hline\hline
improvement (\%) & 2.6 & 3.3 & \textbf{5.4} \\
\hline
\end{tabular}\vspace{-5pt}
\subcaption{\footnotesize Comparing pairwise terms of different methods.}
\end{subtable}

\caption{\footnotesize Ablation study of hyper-parameters.}
\label{tab:ablation}
\end{table}

\subsection{Effectiveness of DPN}
\label{sec:ablation}


All the models evaluated in this section are trained and tested on VOC12.
%

\textbf{Triple Penalty~} The receptive field of b12 indicates the range of triple relations for each pixel.
We examine different settings of the receptive fields, including `10$\times$10', `50$\times$50', and `100$\times$100', as shown in Table \ref{tab:ablation} (a), where
`50$\times$50' achieves the best mIoU, which is sightly better than `100$\times$100'.
For a 512$\times$512 image, this result implies that 50$\times$50 neighborhood is sufficient to capture relations between pixels, while smaller or larger regions tend to under-fit or over-fit the training data.
Moreover, all models of triple relations outperform the `baseline' method that models dense pairwise relations, \ie \vgg+denseCRF \cite{koltun2011efficient}.

\textbf{Label Contexts~} Receptive field of b13 indicates the range of local label context.
To evaluate its effectiveness, we fix the receptive field of b12 as 50$\times$50.
As summarized in Table \ref{tab:ablation} (b), `9$\times$9 mixtures' improves preceding settings by 1.7, 0.5, and 0.2 percent respectively.
We observe large gaps exist between `1$\times$1' and `5$\times$5'.
Note that the 1$\times$1 receptive field of b13 corresponds to learning a global label co-occurrence without considering local spatial contexts.
%
%
Table \ref{tab:ablation} (c) shows that the pairwise terms of DPN are more effective than DSN and DeepLab\footnote{The other deep models such as RNN and Piecewise did not report the exact imrprovements after combining unary and pairwise terms.}.

More importantly, mIoU of all the categories can be improved when increasing the size of receptive field and learning a mixture.
Specifically, for each category, the improvements of the last three settings in Table \ref{tab:ablation} (b) over the first one are 1.2$\pm$0.2, 1.5$\pm$0.2, and 1.7$\pm$0.3, respectively.

\begin{figure}
  \centering
  \includegraphics[width=0.45\textwidth]{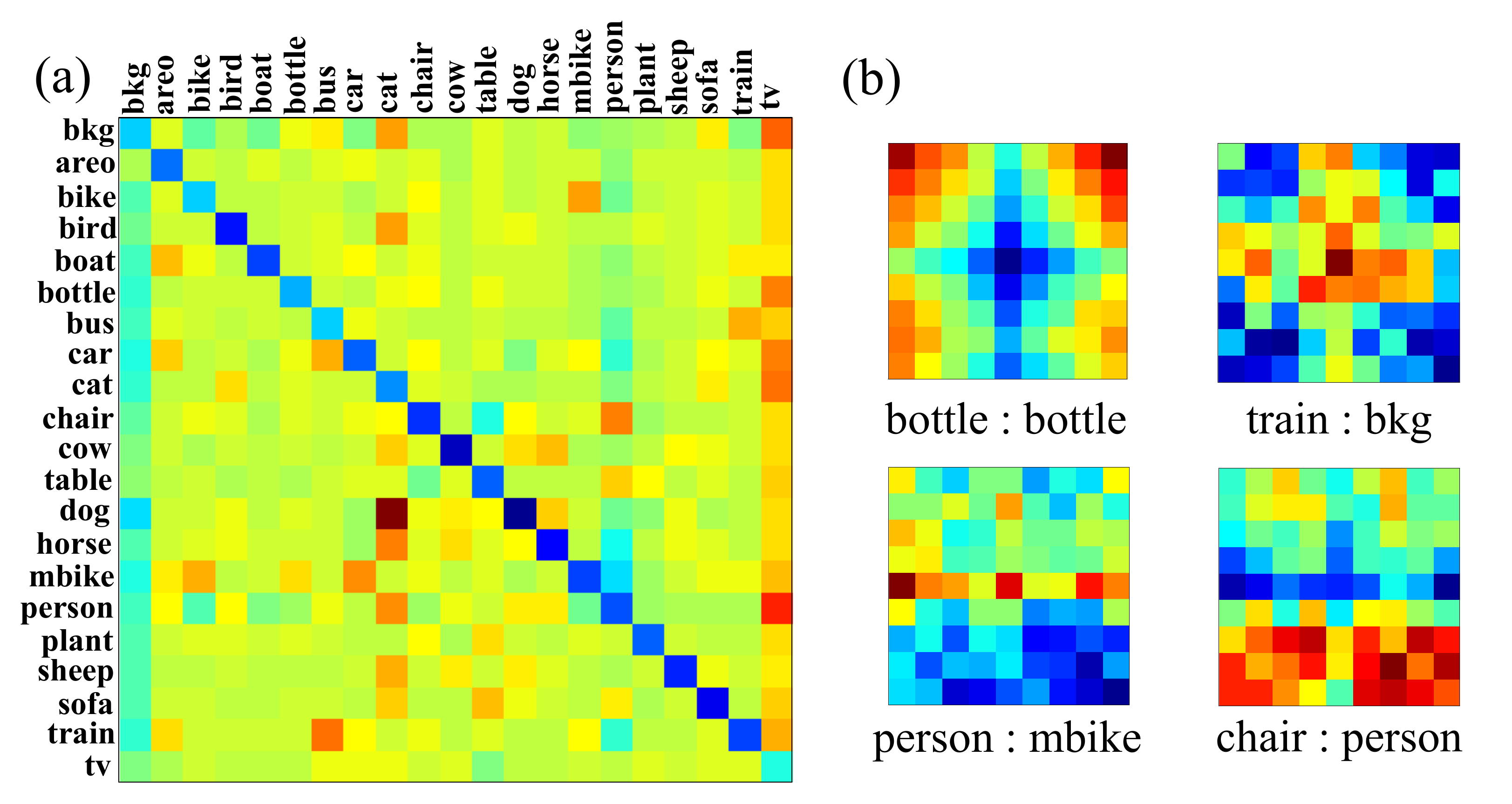}
  \caption{\footnotesize Visualization of (a) learned label compatibility (b) learned contextual information. \textbf{(Best viewed in color)}}
  \label{fig:label}
\end{figure}

We also visualize the learned label compatibilities and contexts in Fig.\ref{fig:label} (a) and (b), respectively.
(a) is obtained by summing each filter in b13 over 9$\times$9 region, indicating how likely a column object would present when a row object is presented. Blue represents high possibility.
(a) is non-symmetry.
For example, when `horse' is presented, `person' is more likely to present than the other objects. Also, `chair' is compatible with `table' and `bkg' is compatible with all the objects.
(b) visualizes some contextual patterns, where `A:B' indicates that when `A' is presented, where `B' is more likely to present.
For example, `bkg' is around `train', `motor bike' is below `person', and `person' is sitting on `chair'.

\begin{figure}
  \centering
  \includegraphics[width=0.45\textwidth]{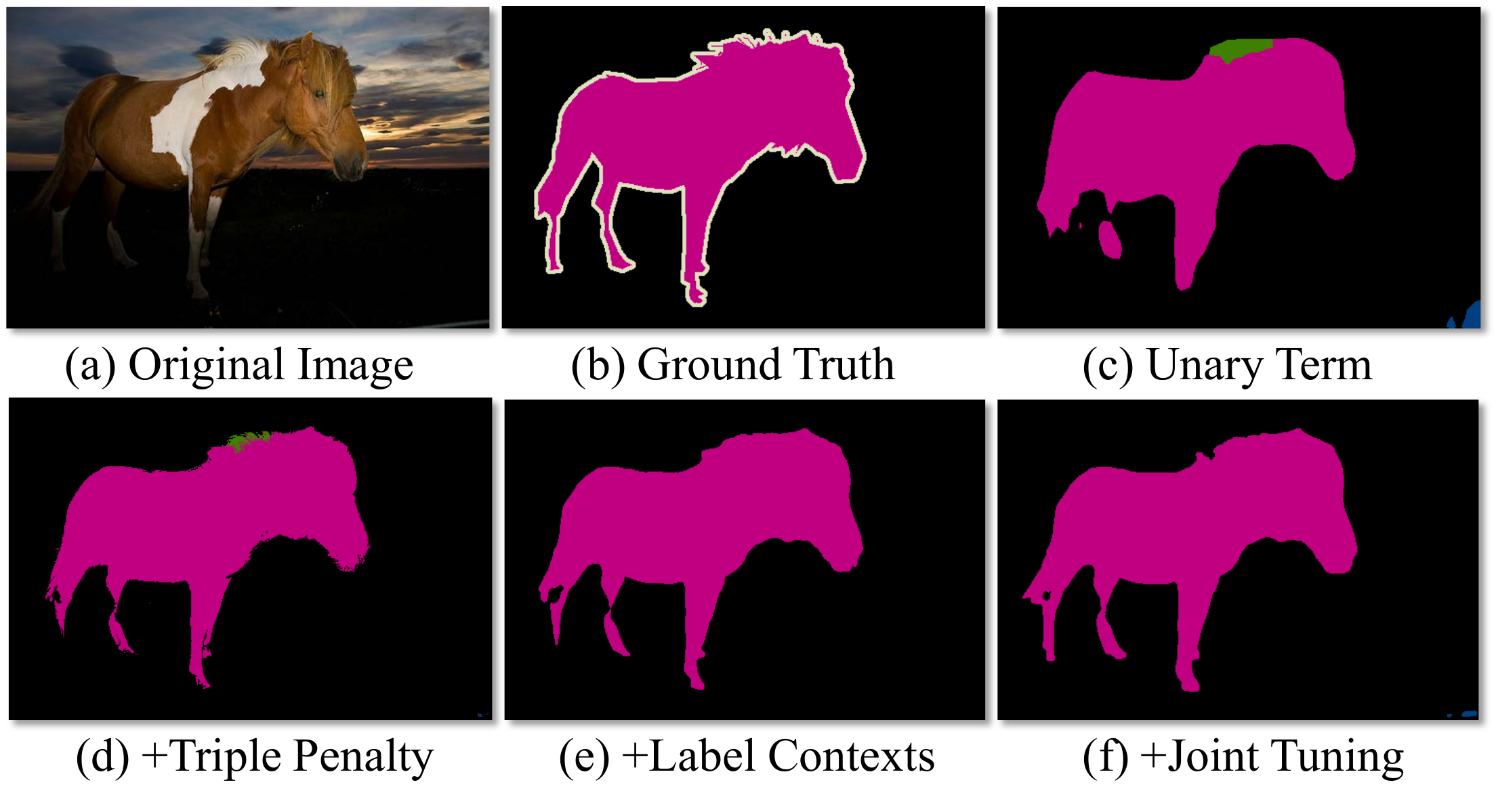}
  \caption{\footnotesize Step-by-step visualization of DPN. \textbf{(Best viewed in color)}}
  \label{fig:vis_pipeline}
\end{figure}

\textbf{Incremental Learning} As discussed in Sec.\ref{sec:learn}, DPN is trained in an incremental manner.
The right hand side of Table \ref{tab:perclass} (a) demonstrates that each stage leads to performance gain compared to its previous stage.
For instance, `triple penalty' improves `unary term' by 2.3 percent, while `label contexts' improves `triple penalty' by 1.8 percent.
More importantly, joint fine-tuning all the components (\ie unary terms and pairwise terms) in DPN achieves another gain of 1.3 percent.
A step-by-step visualization is provided in Fig.\ref{fig:vis_pipeline}.

\begin{figure}
  \centering
  \includegraphics[width=0.45\textwidth]{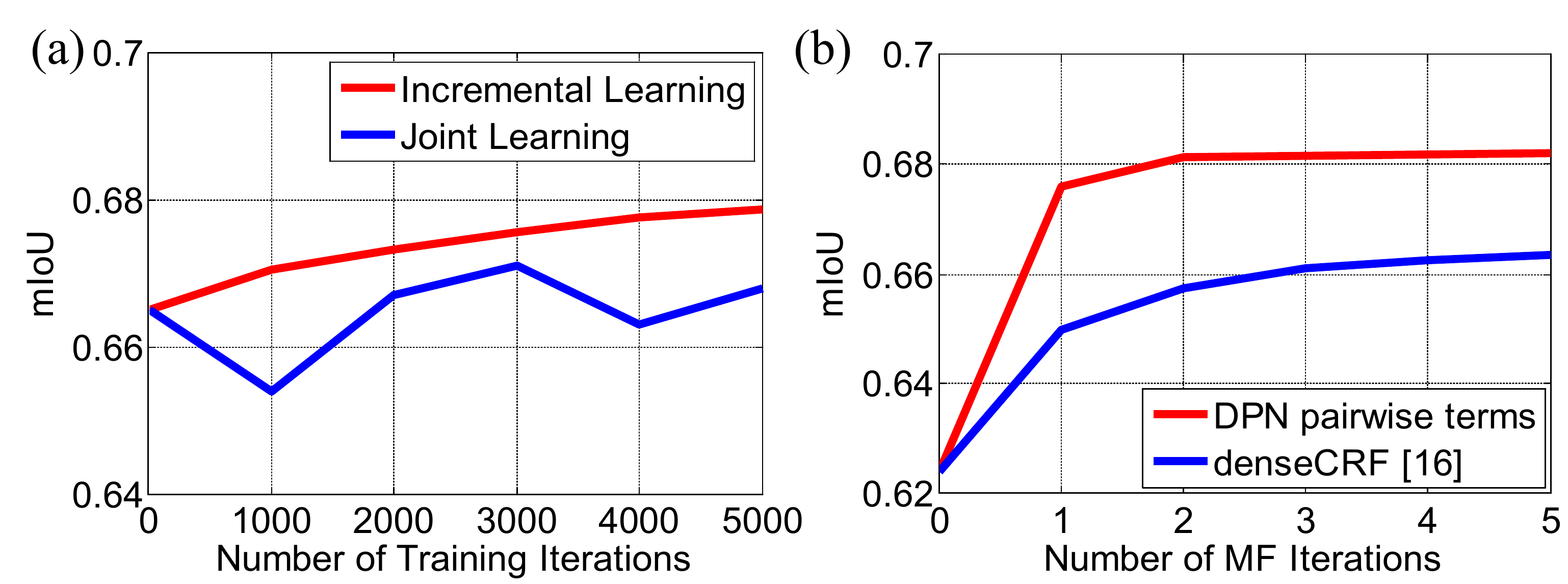}
  \caption{\footnotesize Ablation study of (a) training strategy (b) required MF iterations. \textbf{(Best viewed in color)}}
  \label{fig:strategy}
\end{figure}

We also compare `incremental learning' with `joint learning', which fine-tunes all the components of DPN at the same time.
The training curves of them are plotted in Fig.\ref{fig:strategy} (a), showing that the former leads to higher and more stable accuracies with respect to different iterations, while the latter may get stuck at local minima.
This difference is easy to understand, because incremental learning only introduces new parameters until all existing parameters have been fine-tuned.

\textbf{One-iteration MF} DPN approximates one iteration of MF.
Fig.\ref{fig:strategy} (b) illustrates that DPN reaches a good accuracy with one MF iteration.
%
%
A CRF \cite{koltun2011efficient} with dense pairwise edges needs more than 5 iterations to converge. It also has a large gap compared to DPN.
%
Note that the existing deep models such as \cite{chen2014semantic, zheng2015conditional, schwing2015fully} required 5$\sim$10 iterations to converge as well.

\begin{figure}[t]
  \centering
  \includegraphics[width=0.45\textwidth]{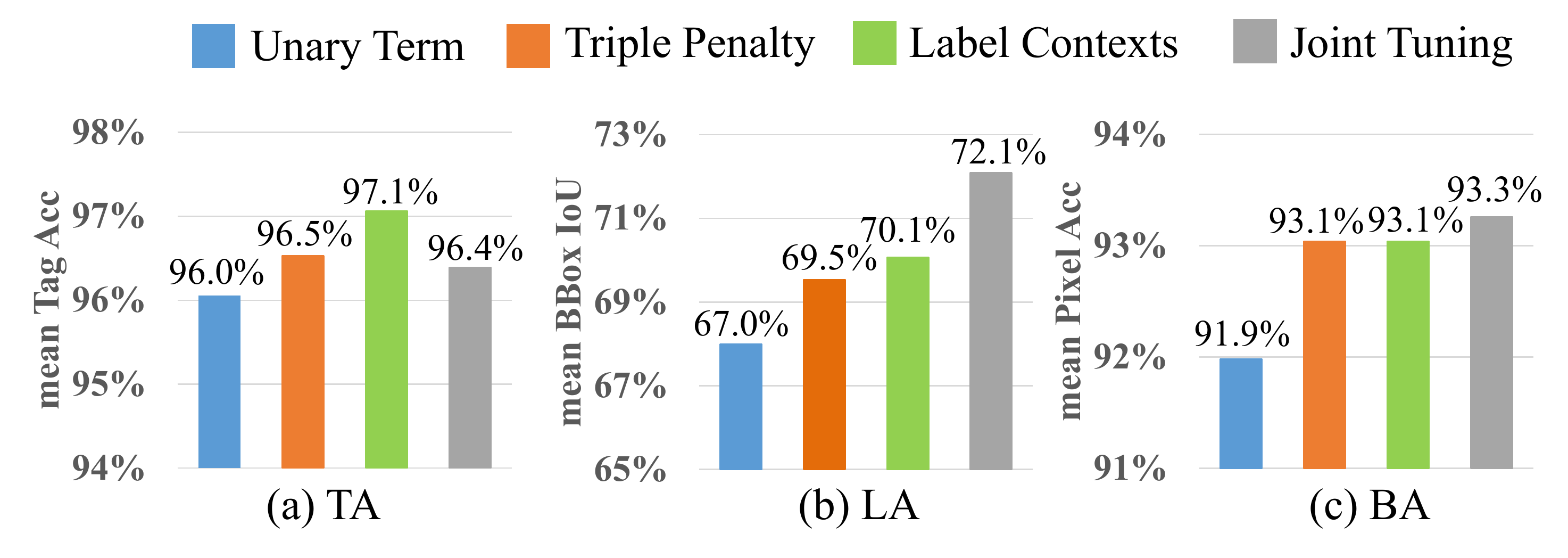}
  \caption{\footnotesize Stage-wise analysis of (a) mean tagging accuracy (b) mean localization accuracy (c) mean boundary accuracy.}
  \label{fig:error}
\end{figure}

\begin{table*}
\scriptsize

\begin{subtable}{\linewidth}
\vspace{4pt}
\centering
\begin{tabular}{l|p{8pt}p{8pt}p{8pt}p{8pt}p{8pt}p{8pt}p{8pt}p{8pt}p{8pt}p{8pt}p{8pt}p{8pt}p{8pt}p{8pt}p{8pt}p{8pt}p{8pt}p{8pt}p{8pt}p{8pt}|p{12pt}}
\hline
 & areo & bike & bird & boat & bottle & bus & car & cat & chair & cow & table & dog & horse & mbike & person & plant & sheep & sofa & train & tv & \textbf{Avg.} \\
\hline\hline
Unary Term (mIoU) & 77.5 & 34.1 & 76.2 & 58.3 & 63.3 & 78.1 & 72.5 & 76.5 & 26.6 & 59.9 & 40.8 & 70.0 & 62.9 & 69.3 & 76.3 & 39.2 & 70.4 & 37.6 & 72.5 & 57.3 & 62.4 \\
~~+ Triple Penalty & 82.3 & 35.9 & 80.6 & 60.1 & 64.8 & 79.5 & 74.1 & 80.9 & 27.9 & 63.5 & 40.4 & 73.8 & 66.7 & 70.8 & 79.0 & 42.0 & 74.1 & 39.1 & 73.2 & 58.5 & 64.7 \\
~~+ Label Contexts & 83.2 & 35.6 & \textbf{82.6} & 61.6 & 65.5 & 80.5 & 74.3 & \textbf{82.6} & 29.9 & \textbf{67.9} & 47.5 & 75.2 & \textbf{70.3} & 71.4 & 79.6 & 42.7 & \textbf{77.8} & 40.6 & 75.3 & 59.1 & 66.5 \\
~~+ Joint Tuning & \textbf{84.8} & \textbf{37.5} & 80.7 & \textbf{66.3} & \textbf{67.5} & \textbf{84.2} & \textbf{76.4} & 81.5 & \textbf{33.8} & 65.8 & \textbf{50.4} & \textbf{76.8} & 67.1 & \textbf{74.9} & \textbf{81.1} & \textbf{48.3} & 75.9 & \textbf{41.8} & \textbf{76.6} & \textbf{60.4} & \textbf{67.8} \\
\hline\hline
TA (tagging Acc.) & 98.8 & 97.9 & 98.4 & 97.7 & 96.1 & 98.6 & 95.2 & 96.8 & 90.1 & 97.5 & 95.7 & 96.7 & 96.3 & 98.1 & 93.3 & 96.1 & 98.7 & 92.2 & 97.4 & 96.3 & 96.4 \\
LA (bIoU)  & 81.7 & 76.3 & 75.5 & 70.3 & 54.4 & 86.4 & 70.6 & 85.6 & 51.8 & 79.6 & 57.1 & 83.3 & 79.2 & 80.0 & 74.1 & 53.1 & 79.1 & 68.4 & 76.3 & 58.8 & 72.1\\
BA (boundary Acc.) & 95.9 & 83.9 & 96.9 & 92.6 & 93.8 & 94.0 & 95.7 & 95.6 & 89.5 & 93.3 & 91.4 & 95.2 & 94.2 & 92.7 & 94.5 & 90.4 & 94.8 & 90.5 & 93.7 & 96.6 & 93.3 \\
\hline
\end{tabular}\vspace{-5pt}
\subcaption{\footnotesize Per-class results on VOC12 {\em val}.}
\end{subtable}

\begin{subtable}{\linewidth}
\vspace{4pt}
\centering
\begin{tabular}{l|p{8pt}p{8pt}p{8pt}p{8pt}p{8pt}p{8pt}p{8pt}p{8pt}p{8pt}p{8pt}p{8pt}p{8pt}p{8pt}p{8pt}p{8pt}p{8pt}p{8pt}p{8pt}p{8pt}p{8pt}|p{12pt}}
\hline
 & areo & bike & bird & boat & bottle & bus & car & cat & chair & cow & table & dog & horse & mbike & person & plant & sheep & sofa & train & tv & mIoU \\
\hline\hline
FCN \cite{long2014fully} & 76.8 & 34.2 & 68.9 & 49.4 & 60.3 & 75.3 & 74.7 & 77.6 & 21.4 & 62.5 & 46.8 & 71.8 & 63.9 & 76.5 & 73.9 & 45.2 & 72.4 & 37.4 & 70.9 & 55.1 & 62.2 \\
Zoom-out \cite{mostajabi2014feedforward} & 85.6 & 37.3 & 83.2 & 62.5 & 66.0 & 85.1 & 80.7 & 84.9 & 27.2 & 73.2 & 57.5 & 78.1 & 79.2 & 81.1 & 77.1 & 53.6 & 74.0 & 49.2 & 71.7 & 63.3 & 69.6 \\
Piecewise \cite{lin2015efficient} & 87.5 & 37.7 & 75.8 & 57.4 & 72.3 & 88.4 & 82.6 & 80.0 & 33.4 & 71.5 & 55.0 & 79.3 & 78.4 & 81.3 & 82.7 & 56.1 & 79.8 & 48.6 & 77.1 & 66.3 & 70.7 \\
DeepLab \cite{chen2014semantic} & 84.4 & 54.5 & 81.5 & 63.6 & 65.9 & 85.1 & 79.1 & 83.4 & 30.7 & 74.1 & 59.8 & 79.0 & 76.1 & 83.2 & 80.8 & 59.7 & 82.2 & 50.4 & 73.1 & 63.7 & 71.6 \\
RNN \cite{zheng2015conditional} & 87.5 & 39.0 & 79.7 & 64.2 & 68.3 & 87.6 & 80.8 & 84.4 & 30.4 & 78.2 & 60.4 & 80.5 & 77.8 & 83.1 & 80.6 & 59.5 & 82.8 & 47.8 & 78.3 & 67.1 & 72.0 \\
\hline
WSSL$^\dagger$ \cite{papandreou2015weakly} & 89.2 & 46.7 & 88.5 & 63.5 & 68.4 & 87.0 & 81.2 & 86.3 & 32.6 & 80.7 & 62.4 & 81.0 & 81.3 & 84.3 & 82.1 & 56.2 & 84.6 & 58.3 & 76.2 & 67.2 & 73.9 \\
RNN$^\dagger$ \cite{zheng2015conditional} & 90.4 & 55.3 & 88.7 & 68.4 & 69.8 & 88.3 & 82.4 & 85.1 & 32.6 & 78.5 & 64.4 & 79.6 & 81.9 & \textbf{86.4} & 81.8 & 58.6 & 82.4 & 53.5 & 77.4 & 70.1 & 74.7 \\
BoxSup$^\dagger$ \cite{dai2015boxsup} & \textbf{89.8} & 38.0 & \textbf{89.2} & \textbf{68.9} & 68.0 & 89.6 & 83.0 & \textbf{87.7} & 34.4 & 83.6 & \textbf{67.1} & 81.5 & 83.7 & 85.2 & 83.5 & 58.6 & 84.9 & 55.8 & \textbf{81.2} & \textbf{70.7} & 75.2 \\
\hline\hline
DPN & 87.7 & 59.4 & 78.4 & 64.9 & 70.3 & 89.3 & 83.5 & 86.1 & 31.7 & 79.9 & 62.6 & 81.9 & 80.0 & 83.5 & 82.3 & 60.5 & 83.2 & 53.4 & 77.9 & 65.0 & 74.1 \\
DPN$^\dagger$ & 89.0 & \textbf{61.6} & 87.7 & 66.8 & \textbf{74.7} & \textbf{91.2} & \textbf{84.3} & 87.6 & \textbf{36.5} & \textbf{86.3} & 66.1 & \textbf{84.4} & \textbf{87.8} & 85.6 & \textbf{85.4} & \textbf{63.6} & \textbf{87.3} & \textbf{61.3} & 79.4 & 66.4 & \textbf{77.5} \\
\hline
\end{tabular}\vspace{-5pt}

\subcaption{\footnotesize Per-class results on VOC12 {\em test}. The approaches pre-trained on COCO \cite{lin2014microsoft} are marked with $^\dagger$.}

\end{subtable}

\caption{\footnotesize Per-class results on VOC12.}
\label{tab:perclass}

\end{table*}

\textbf{Different Components Modeling Different Information~}
We further evaluate DPN using three metrics.
The results are given in Fig.\ref{fig:error}.
For example, (a) illustrates that the tagging accuracy can be improved in the third stage, as it captures label co-occurrence with a mixture of contextual patterns. However, TA is decreased a little after the final stage.
%
Since joint tuning maximizes segmentation accuracies by optimizing all components together,
extremely small objects, which rarely occur in VOC training set, are discarded.
%
%
%
As shown in (b), accuracies of object localization are significantly improved in the second and the final stages.
This is intuitive because the unary prediction can be refined by long-range and high-order pixel relations, and joint training further improves results.
(c) discloses that the second stage also captures object boundary, since it measures dissimilarities between pixels.

\textbf{Per-class Analysis~}
Table \ref{tab:perclass} (a) reports the per-class accuracies of four evaluation metrics, where the first four rows represent the mIoU of four stages, while the last three rows represent TA, LA, and BA, respectively.
We have several valuable observations, which motivate future researches.
(1) Joint training benefits most of the categories, except animals such as `bird', `cat', and `cow'.
Some instances of these categories are extremely small so that joint training discards them for smoother results.
(2) Training DPN with pixelwise label maps implicitly models image-level tags, since it achieves a high averaged TA of 96.4\%.
(3) Object localization always helps.
However, for the object with complex boundary such as `bike', its mIoU is low even it can be localized, \eg `bike' has high LA but low BA and mIoU.
%
%
(4) Failures of different categories have different factors.
With these three metrics, they can be easily identified.
%
For example, the failures of  `chair', `table', and `plant'  are caused by the difficulties to accurately capture their bounding boxes and boundaries.
%
Although `bottle' and `tv' are also difficult to localize, they achieve moderate mIoU because of their regular shapes.
In other words, mIoU of `bottle' and `tv' can be significantly improved if they can be accurately localized.

\subsection{Overall Performance}\label{sec:overall}


As shown in Table \ref{tab:perclass} (b), we compare DPN with the best-performing methods\footnote{The results of these methods were presented in either the published papers or arXiv pre-prints.} on VOC12 test set based on two settings, \ie with and without pre-training on COCO.
%
The approaches pre-trained on COCO are marked with `$\dag$'.
We evaluate DPN on several scales of the images and then average the results following \cite{chen2014semantic, lin2015efficient}.
%

DPN outperforms all the existing methods that were trained on VOC12, but DPN needs only one MF iteration to solve MRF, other than 10 iterations of RNN, DeepLab, and Piecewise.
By averaging the results of two DPNs,
we achieve 74.1\% accuracy on VOC12 without outside training data.
%
As discussed in Sec.\ref{sec:learn}, MF iteration is the most complex step even when it is implemented as convolutions.
Therefore, DPN at least reduces 10$\times$ runtime compared to previous works.

Following \cite{zheng2015conditional,dai2015boxsup}, we pre-train DPN with COCO, where 20 object categories that are also presented in VOC12 are selected for training.
%
%
%
A single DPN$^\dagger$ has achieved 77.5\% mIoU on VOC12 test set.
%
%
As shown in Table \ref{tab:perclass} (b), we observe that DPN$^\dagger$ achieves best performances on more than half of the object classes.
Please refer to the appendices for visual quality comparisons.


\section{Conclusion}

We proposed Deep Parsing Network (DPN) to address semantic image segmentation,
which has several appealing properties.
First, DPN unifies the inference and learning of unary term and pairwise terms in a single convolutional network.
No iterative inference are required during back-propagation.
Second, high-order relations and mixtures of label contexts are incorporated to its pairwise terms modeling,
making existing works serve as special cases.
Third, DPN is built upon conventional operations of CNN, thus easy to be parallelized and speeded up.

DPN achieves state-of-the-art performance on VOC12, and multiple valuable facts about semantic image segmention are revealed through extensive experiments.
%
Future directions include investigating the generalizability of DPN to more challenging scenarios,
\eg large number of object classes and substantial appearance/scale variations.




{\small
\bibliographystyle{ieee}
\bibliography{egbib}
}

\vspace{40pt}

\begin{appendix}

\textbf{\Large Appendices}

\vspace{20pt}

\section{Fast Implementation of Locally Convolution}

\textbf{b12} in DPN is a locally convolutional layer.
As mentioned in Eqn.(\ref{eq:old_p}), the local filters in \textbf{b12} are computed by the distances between RGB values of the pixels.
XY coordinates are omitted here because they could be pre-computed.
To accelerate the computation of locally convolution, lookup table-based filtering approach is employed.
We first construct a lookup table storing distances between any two pixel intensities (ranging from 0 to 255),
which results in a $256 \times 256$ matrix.
Then when we perform locally convolution, the kernels' coefficients can be obtained efficiently by just looking up the table.

\section{Visual Quality Comparisons}

\begin{figure*}[t]
  \centering
  \includegraphics[width=0.9\textwidth]{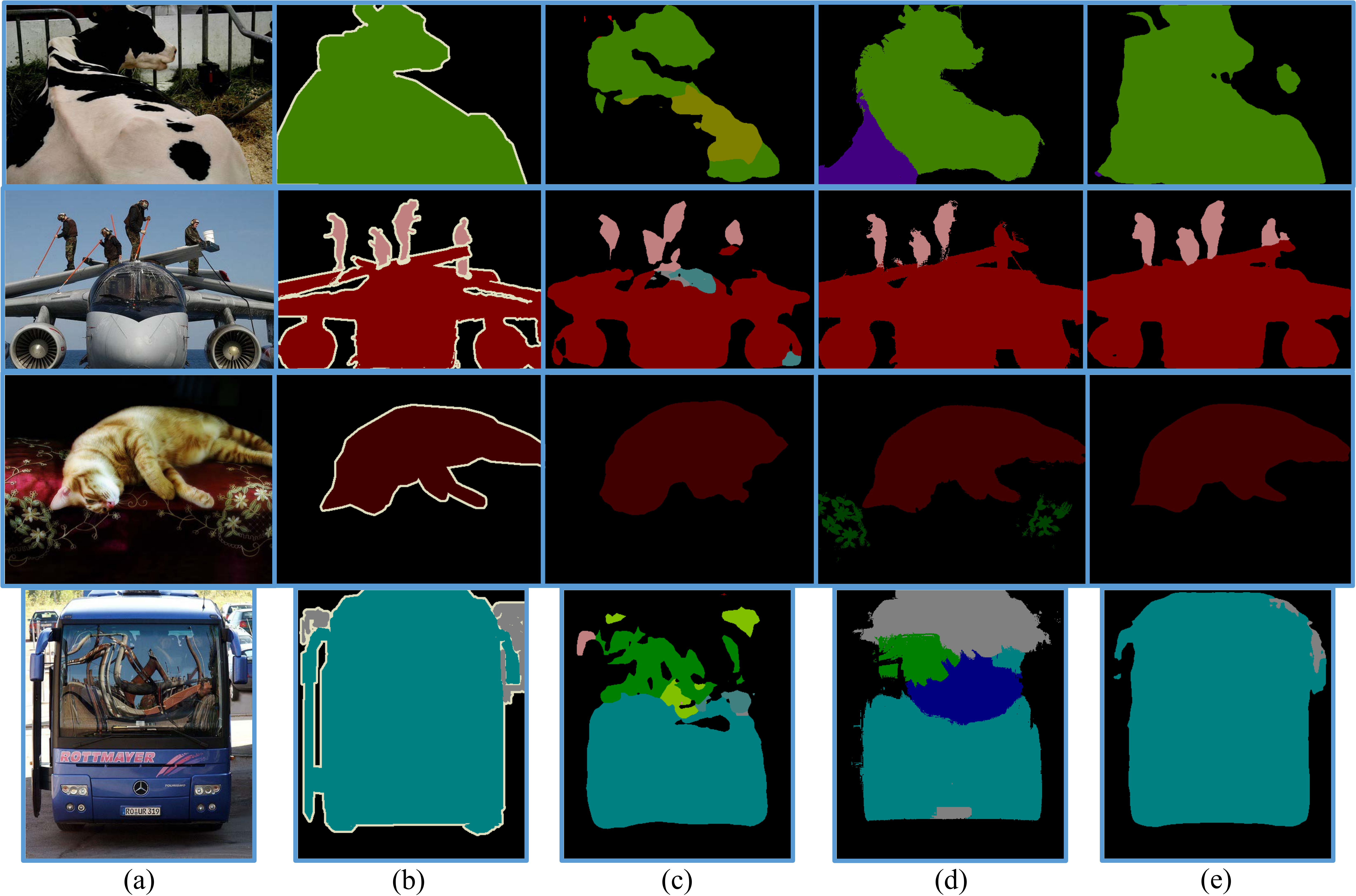}
  \caption{Visual quality comparison of different semantic image segmentation methods: (a) input image (b) ground truth (c) FCN \cite{long2014fully} (d) DeepLab \cite{chen2014semantic} and (e) DPN.}
  \label{fig:visualquality_comparison}
\end{figure*}

In the following, we inspect visual quality of obtained label maps.
Fig.\ref{fig:visualquality_comparison} demonstrates the comparisons of DPN with FCN \cite{long2014fully} and DeepLab \cite{chen2014semantic}.
We use the publicly released model\footnote{\url{http://dl.caffe.berkeleyvision.org/fcn-8s-pascal.caffemodel}} to re-generate label maps of FCN while the results of DeepLab are extracted from their published paper.
DPN generally makes more accurate predictions in both image-level and instance-level.

\begin{figure*}
  \centering
  \includegraphics[width=0.9\textwidth]{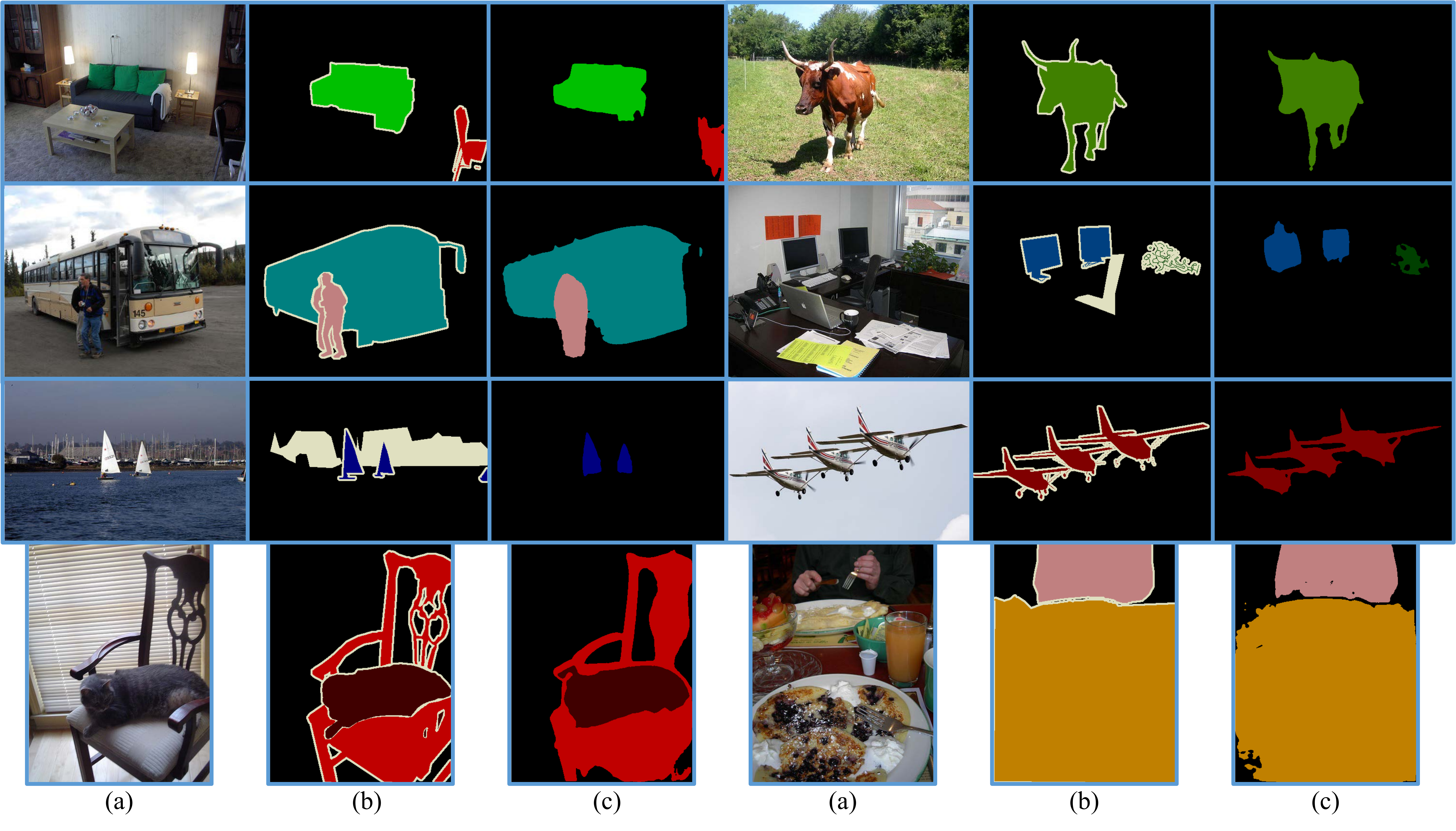}
  \caption{Visual quality of DPN label maps: (a) input image (b) ground truth (white labels indicating ambiguous regions) and (c) DPN.}
  \label{fig:visualquality_self}
\end{figure*}

We also include more examples of DPN label maps in Fig.\ref{fig:visualquality_self}.
We observe that learning local label contexts helps differentiate confusing objects and learning triple penalty facilitates the capturing of intrinsic object boundaries.


\end{appendix}

\end{document}